\documentclass{article}

\usepackage{spconf,amsmath,graphicx}
\usepackage{amssymb}
\usepackage{amsthm}
\usepackage{bm}
\usepackage{array}
\usepackage{float}
\usepackage{breqn}
\usepackage{multirow}

\usepackage{bm}
\usepackage{amssymb,amsmath,graphicx,bbm,color,booktabs}
\usepackage{amsopn}

\newcommand{\vw}{\bm{w}}

\newcommand{\vphi}    {\bm{\phi}}

  \newcommand{\Sc}{\mathcal{S}}

    \newcommand{\Yc}{\mathcal{Y}}

\newcommand{\R}{\mathbb{R}}

\newcommand{\reals}{\mathbb{R}}

\DeclareMathOperator*{\argmin}{argmin}
\DeclareMathOperator*{\argmax}{argmax}

\newcommand{\eq}[1]{(\protect\ref{#1})}

\newcommand{\sx}{\bar{\x}}

\newcommand{\sy}{\bar{\y}}
\newcommand{\syh}{\sy'}

\renewcommand{\eqref}[1]{Eq.~(\ref{#1})}

\newcommand{\segmentor}{DeepSegmentor }

\def \x{{\mathbf x}}
\def \y{{\mathbf y}}

\title{Sequence Segmentation using joint RNN and \\
		structured prediction models}

\name{Yossi Adi$^1$, Joseph Keshet$^{1}$\thanks{Supported in part by NIH grant 1R21HD077140.}, Emily Cibelli$^2$, Matthew Goldrick$^2$}
\address{$^1$Department of Computer Science, Bar-Ilan University, Ramat-Gan, Israel\\
$^2$Department of Linguistics, Northwestern University, Evanston, IL, USA}

\begin{document}
%
\maketitle
\begin{abstract}
We describe and analyze a simple and effective algorithm for sequence segmentation applied to speech processing tasks. We propose a neural architecture that is composed of two modules trained jointly: a recurrent neural network (RNN) module and a structured prediction model. The RNN outputs are considered as feature functions to the structured model. The overall model is trained with a structured loss function which can be designed to the given segmentation task. We demonstrate the effectiveness of our method by applying it to two simple tasks commonly used in phonetic studies: word segmentation and voice onset time segmentation. Results suggest the proposed model is superior to previous methods, obtaining state-of-the-art results on the tested datasets.
\end{abstract}
\begin{keywords}
Sequence segmentation,  recurrent neural networks (RNNs), structured prediction, word segmentation, voice onset time
\end{keywords}

\label{sec:intro}
\section{Introduction}

Sequence segmentation is an important task for many speech and audio applications such as speaker diarization, laboratory phonology research, speech synthesis, and automatic speech recognition (ASR). Segmentation models can be used as a pre-process step to clean the data (e.g., removing non-speech regions such as music or noise to reduce ASR error \cite{kubala1996transcribing, rybach2009audio}). They can also be used as tools in clinically- or theoretically-focused phonetic studies that utilize acoustic properties as a dependent measure. For example, voice onset time, a key feature distinguishing voiced and voiceless consonants across languages \cite{lisker1964cross}, is important both in ASR \cite{hansen2010automatic-short}, clinical \cite{auzou2000voice-short}, and theoretical studies \cite{paterson2011interactions}. 

Previous work on speech sequence segmentation focuses on generative models such as hidden Markov models (see for example \cite{toledano2003automatic} and the references therein); on discriminative methods \cite{rybach2009audio, keshet2007large, sonderegger2012automatic}; or on deep learning \cite{adi2015vowel, adiautomatic}. 

Inspired by the recent work on combined deep network and structured prediction models \cite{do2010neural, zheng2015conditional, chen2015learning, kiperwasser2016simple, lample2016neural}, we would like to further improve performance on speech sequence segmentation and propose a new efficient joint deep network and structure prediction model. Specifically, we jointly optimize RNN and structured loss parameters by using RNN outputs as feature functions for a structured prediction model. First, an RNN encodes the entire speech utterance and outputs new representation for each of the frames. Then, an efficient search is applied over all possible segments so that the most probable one can be selected. We evaluate this approach using two tasks: word segmentation and voice onset time segmentation. In both  tasks the input is a speech segment and the goal is to determine the boundaries of the defined event. We show that the proposed approach outperforms previous methods on these two segmentation tasks.

\label{sec:prob_def}
\section{Problem Setting}

In the problem of speech segmentation we are provided with a speech utterance, denoted as $\sx = (\x_1,\ldots,\x_T)$, represented as a sequence of acoustic feature vectors, where each $\x_t\in\R^D$ $(1\leq t \leq T)$ is a $D$-dimensional vector. The length of the speech utterance, $T$, is not a fixed value, since the input utterances can have different durations. 

Each input utterance is associated with a timing sequence, denoted by $\sy = (y_1,\ldots,y_p)$, where $p$ can vary across different inputs. Each element $y_i \in \Yc$, where $\Yc=\{1,\ldots,T\}$  indicates the start time of a new event in the speech signal. We annotate all the possible timing sequence of size $p$ by $\Yc^p$ 

For example, in \emph{word segmentation} the goal is to segment a word from silence and noise in the signal. In this case the size of $\sy$ is 2, namely word onset and offset. However, in phoneme segmentation the goal is to segment every phoneme in a spoken word. In this case the size of $\sy$ is different for each input sequence. 

Generally, our method is suitable for different sequence size $|\sy|$. In this paper we focused on $|\sy|=2$,  and leave the problem of $|\sy| > 2$ to future work.

\label{sec:model}
\section{Model Description}

We now describe our model in greater detail. First, we present the  structured prediction framework and then discuss how it is combined with an RNN.

\subsection{Structured Prediction}

We consider the following prediction rule with $\vw \in \reals^d$, such that $\syh_{\vw}$ is a good approximation to the true label of $\sx$, as follows:
\begin{dmath}
\label{eq:yw}
\syh_{\vw}(\sx) = \argmax_{\sy \in \Yc} ~ \vw^\top \vphi(\sx, \sy)
\end{dmath} 

Following the structured prediction framework, we assume there exists some unknown probability distribution $\rho$ over pairs $(\sx,\sy)$ where $\sy$ is the desired output (or reference output) for input $\sx$. Both $\sx$ and $\sy$ are usually structured objects such as sequences, trees, etc. Our goal is to set $\vw$ so as to minimize the expected cost, or the \emph{risk},
\begin{equation}
	\label{eq:w*}
	\vw^* = \argmin_{\vw} ~ \mathbb{E}_{(\sx,\sy) \sim \rho} [\ell(\sy,\syh_{\vw}(\sx))]. 
\end{equation}
This objective function is hard to minimize directly since the distribution $\rho$ is unknown. We use a training set $\Sc = \{(\sx_1,\sy_1),\ldots,(\sx_m,\sy_m)\}$ of $m$ examples that are drawn i.i.d. from $\rho$, and replace the expectation in \eq{eq:w*} with a mean over the training set.

The cost is often a combinatorial non-convex quantity, which is hard to minimize. Hence, instead of minimizing the cost directly, we minimize a slightly different function called a \emph{surrogate loss}, denoted $\bar{\ell}(\vw,\sx,\sy)$, and closely related to the cost. Overall, the objective function in \eq{eq:w*} transforms into the following objective function, denoted as $F$: 
\begin{dmath}
\label{eq:reg-loss}
F(\vw, \sx, \sy) = \frac{1}{m}\sum_{i=1}^{m} \bar{\ell}(\vw, \sx, \sy)
\end{dmath} 

In this work the surrogate loss function is the structural hinge loss \cite{tsochantaridis2005large} defined as 
\[ 
\bar{\ell}(\vw, \sx, \sy) = \max_{\syh \in \Yc} ~ \left[\ell(\sy,\syh) - \vw^\top\vphi(\sx,\sy) + \vw^\top\vphi(\sx,\syh) \right]
\]

Usually, $\vphi(\sx, \sy)$ is manually chosen using data analysis techniques and involves manipulation on local and global features. In the next subsection we describe how to use an RNN as feature functions.

\subsection{Recurrent Neural Networks as Feature Functions}
RNN is a deep network architecture that can model the behavior of dynamic temporal sequences using an internal state which can be thought of as memory \cite{elman1991distributed, graves2013speech}. RNN provides the ability to predict the current frame label based on the previous frames.  Bidirectional RNN is a model composed of two RNNs: the first is a standard RNN while the second reads the input backwards. Such a model can predict the current frame based on both past and future frames. By using the RNN outputs we can jointly train the structured and network models.

\begin{figure}[htb]
\centering   
\includegraphics[scale=0.6]{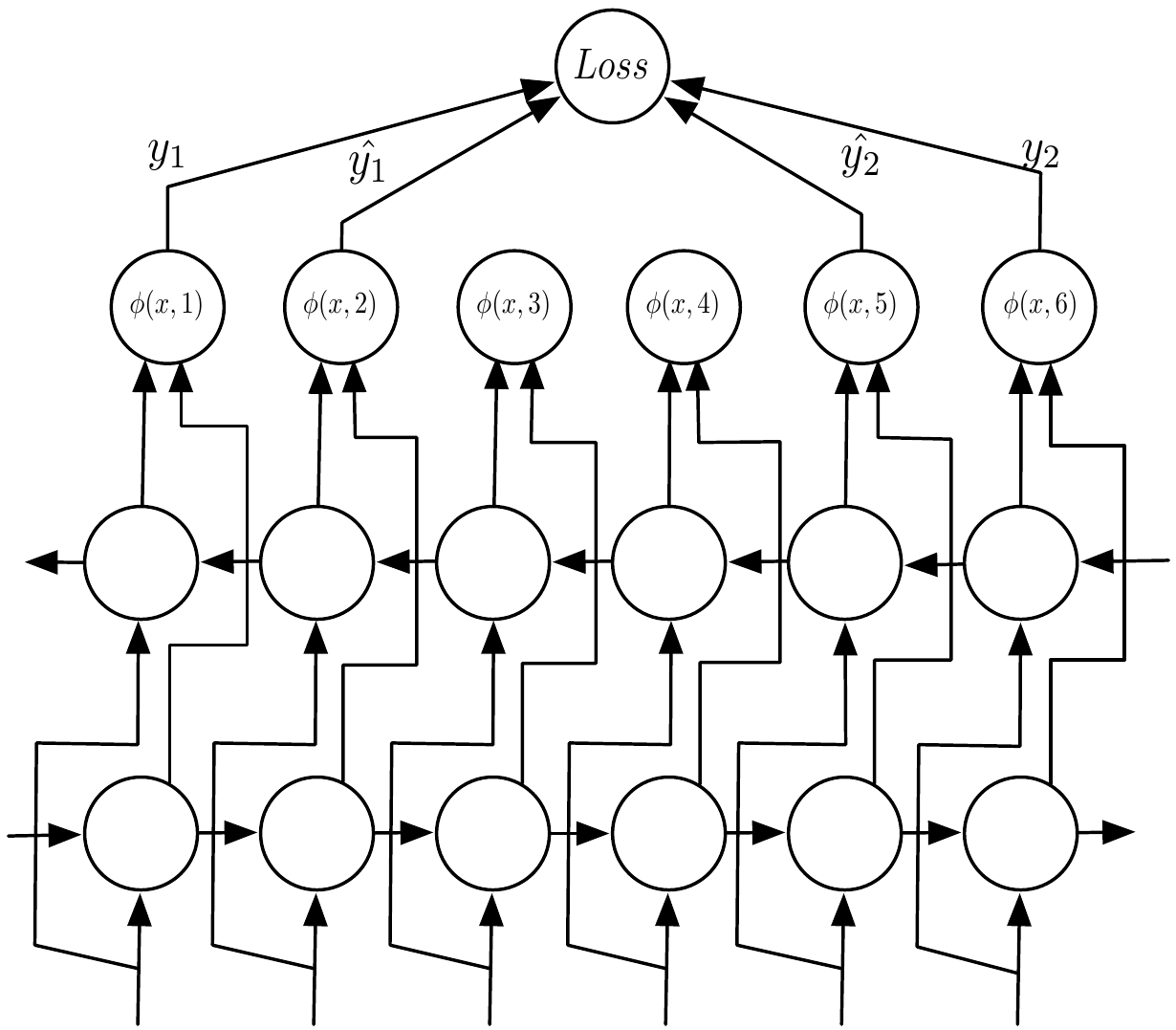}
\caption{An illustration for using BI-RNN as feature functions. We search through all possible locations and predict the one with the highest score. In this example the target timing sequence is $(1, 6)$ and the predicted timing sequence is $(2, 5)$.}
 \label{fig:rnn}
\end{figure}

Recall our prediction rule in \eqref{eq:yw}: notice that $\vphi(\sx, \sy)$ can be viewed as $\sum_{i=1}^{p} \vphi'(\sx, y_i)$ where each $\vphi$ can be extracted using different techniques, e.g., hand-crafted, feed-forward neural network, RNNs, etc. We can formulate the prediction rule as follows:
\begin{dmath}
\label{eq:dec_phi}
\syh_{\vw}(\sx) 
= \argmax_{\sy \in \Yc^p} ~ \vw^\top \vphi(\sx, \sy) 
= \argmax_{\sy \in \Yc^p} ~ \vw^\top \sum_{i=1}^{p} \vphi'(\sx, y_i) 
= \argmax_{\sy \in \Yc^p} ~ \vw^\top \sum_{i=1}^{p} \text{RNN}(\sx, y_i),
\end{dmath}
where the RNN can be of any type and architecture. For example, we can use bidirectional RNN and consider $\vphi$ as the concatenation of both outputs $\text{BI-RNN}_{\text{forward}}\oplus \text{BI-RNN}_{\text{backward}}$. This is depicted in Figure~\ref{fig:rnn}. We call our model \emph{\segmentor}.

Our goal is to find the model parameters so as to minimize the risk as in \eqref{eq:w*}. Recall, we use the structural hinge loss function, and since both the loss function and the RNN are differentiable we can optimize them using gradient based methods such as stochastic gradient descent (SGD). In order to optimize the network parameters using the back-propagation algorithm \cite{rumelhart1988learning}, we must find the outer derivative of each layer with respect to the model parameters and inputs.

The derivative of the loss layer with respect to the layer parameters $\vw$ for the training example $(\sx, \sy)$ is
$$
\frac{\partial{F}}{\partial{\vw}} = \vphi(\sx,\sy^\ell_{\vw}) - \vphi(\sx, \sy),
$$
where
\begin{equation}\label{eq:loss_adjusted}
\sy^{\ell}_{\vw} = \argmax_{\sy' \in \Yc^p} ~ \vw^\top \vphi(\sx, \sy') + \ell(\sy,\sy').
\end{equation}
Similarly, the derivatives with respect to the layer's inputs are
$$
\frac{\partial{F}}{\partial{\vphi(\sx, \sy)}} = -\vw ~~~~~~~
\frac{\partial{F}}{\partial{\vphi(\sx, \syh)}} = \vw.
$$
The derivatives of the rest of the layers are the same as an RNN model. 

\label{sec:res}
\section{Experimental Results}
We investigate two segmentation problems; word segmentation and voice onset time segmentation. We describe each of them in details in the following subsections.\footnote{All models were implemented using Torch7 toolkit \cite{collobert2011torch7,leonard2015rnn}}

\subsection{Word Segmentation}
In the problem of word segmentation we are provided with a speech utterance which contains a single word; our goal is to predict its start and end times. The ability to determine these timings is crucial to phonetic studies that measure speaker properties (e.g. response time \cite{fink2016domain}) or as a preprocessing step for other phonetic analysis tools \cite{adiautomatic, adi2015vowel, sonderegger2012automatic, keshet2007large, rosenfelder2014fave}.

\subsubsection{Dataset}
Our dataset comes from a laboratory study by Fink and Goldrick \cite{fink2016domain}. Native English speakers were shown a set of 90 pictures. Some participants produced the name of the picture (e.g., saying ``cat'', ``chair'') while others performed a semantic classification task (e.g., saying ``natural'', ``man-made''). Productions other than the intended response or disfluencies were excluded. Recordings were randomly assigned to two transcribers who annotated the onset and offset of each word. We analyze a subset of the recordings, including data from 60 participants, evenly distributed across tasks.

\subsubsection{Results}
We compare our model to an RNN that was trained using the Negative-Log-Liklihood (NLL). The NLL model makes a binary decision in every frame to predict whether there is voice activity or not. Recall, our goal is to find the start and end times of the word; in this task, the RNN leaves us with a distribution over all possible onsets. To account for this, we apply a smoothing algorithm and find the most probable pair of timings.

We trained the \segmentor model using the structured loss function as in \eq{eq:loss}, denoted as Combined Duration (CD) loss. The motivation for using this function is due to disparities in the manual annotations, which are common and depend both on human errors and objective difficulties in placing the boundaries. Hence we chose a loss function that takes into account the variations in the annotations.
\begin{equation}
\label{eq:loss}
\gamma (\sy,\syh) =  \big[|y_1 -  y'_1| - \tau]_+ + \big[|y_2 - y'_2| - \tau]_+,
\end{equation} 
where $[\pi]_+=\max\{0,\pi\}$, and $\tau$ is a user defined tolerance parameter. 

We use two layers of bidirectional LSTMs for the \segmentor model with dropout \cite{dropout} after each recurrent layer. We extracted the 13 Mel-Frequency Cepstrum Coefficients (MFCCs), without the deltas, every 10 ms, and use them as inputs to the network. We optimize the networks using AdaGrad \cite{duchi2011adaptive}. All parameters were tuned on a dedicated development set for both of the models. As for the NLL models, we trained 4 different models; LSTM with one and two layers, and bidirectional LSTM with one and two layers, denoted as RNN, 2RNN, BI-RNN and BI-2-RNN, respectively. Table ~\ref{tab:WDLoss} summarizes the results for both models. 

\begin{table}[!h]
\small
\renewcommand{\arraystretch}{1.5}
	\centering
	\caption{\it The mean loss for NLL models and \segmentor for the word segmentation task. Results are reported for the onset, offset and overall CD separately. The loss function was measured using \eq{eq:loss} (with $\tau$=0) in frames of 10ms.} 
	
  \vspace{5pt}
  \label{tab:WDLoss}  
  \begin{tabular}{lccccc}
    \hline
    \hline
    & RNN & 2-RNN & BI-RNN & BI-2-RNN & DeepSeg.  \\ 
    \hline
    Onset &  6.0 &  5.84 &  2.88 &  3.48 & \bf{2.02}\\
    Offset &  9.43 &  8.92 &  4.46 &  \bf{3.75} &  3.96 \\
    CD &  15.42 &  14.76 &  7.35 &  7.24 &  \bf{5.98} \\
    \hline
    \hline    
  \end{tabular}
\end{table}

Besides being efficient and more elegant, \segmentor is superior to the NLL models when measuring \eq{eq:loss}, with the exception of BI-2-RNN, which was slightly better for the offset measurement.

\subsection{VOT Segmentation}
Voice onset time (VOT) is the time between the onset of a stop burst and the onset of voicing. As noted in the introduction, it is widely used in theoretical and clinical studies as well as ASR tasks. In this problem the input is a speech utterance containing a single stop consonant, and the output is the VOT onset and offset times. 

We compared our model to two other methods for VOT measurement. First is the \emph{AutoVOT} algorithm \cite{sonderegger2012automatic}. This algorithm follows the structured prediction approach of linear classifier with hand-crafted features and feature-functions. The second algorithm is the \emph{DeepVOT} algorithm \cite{adiautomatic}. This algorithm uses RNNs with NLL as loss function. Hence, it predicts for each frame whether it is related to the VOT or not. Using the RNN predictions, a dynamic programming algorithm is applied to find the best onset and offset times. Our approach combines both of these methods while jointly training RNN with structured loss function. 

\subsubsection{Datasets}
We use two different datasets. The first one, \textsc{pgwords}, is from a laboratory study by Paterson and Goldrick \cite{paterson2011interactions}. American English monolinguals and Brazilian Portuguese (L1)-English bilinguals (24 participants each) named a set of 144 pictures. Productions other than the intended label as well as those with code-switching or disfluencies were excluded. VOT of remaining words was annotated by one transcriber.

The second dataset, \textsc{bb}, consists of spontaneous speech from the 2008 season of Big Brother UK, a British reality television show \cite{bane2010longitudinal, sonderegger2012automatic}. The speech comes from 4 speakers recorded in the ``diary room,'' an acoustically clean environment. VOTs were manually measured by two transcribers.

\subsubsection{Results --- \textsc{PGWORDS}}
For the \textsc{pgwords} dataset we use two layers of bidirectional LSTMs with dropout after each recurrent layer. We use \eq{eq:loss} as our loss function. The input features are the same as in \cite{sonderegger2012automatic, adiautomatic}; overall we have 63 features per frame. We optimize the networks using AdaGrad optimization. All parameters were tuned on a dedicated development set. Table ~\ref{tab:PURLoss} summarizes the results using the same loss function as in \cite{sonderegger2012automatic}. Results suggests that \segmentor outperforms the AutoVOT model over all tolerance values. However, when comparing to DeepVOT,  the picture is mixed. In the lower tolerance values \segmentor is superior to the DeepVOT while for higher values DeepVOT performs better. We believe these results are due to the DeepVOT being less delicate and solving a much coarser problem than the \segmentor; hence, it performs better when considering high tolerance values. We believe the integration between these two systems, (using DeepVOT as pre-training for the \segmentor), will yield more accurate and robust results. We leave this investigation for future work.

\begin{table}[!h]
\small
\renewcommand{\arraystretch}{1.5}
	\centering
	\caption{\it Proportion of differences between automatic and manual measures falling at or below a given tolerance value (in msec). For example, for \emph{DeepVOT}, the difference between automatic and manual measurements in the test set was 2 msec or less in 53.8\% of examples. These results are for the \textsc{pgwords} dataset.} 
  \vspace{5pt}
  \label{tab:PURLoss}  
  \begin{tabular}{lcccccc}
    \hline
    \hline
    Model & $t\le$2 & $t\le$5 & $t\le$10 & $t\le$15 & $t\le$25 & $t\le$50\\
    \hline
	AutoVOT & 49.1 & 81.3 & 93.9 & 96.0 & 97.2 & 98.1\\
    \hline	
    DeepVOT & 53.8 & 91.6 & \bf{97.6} & \bf{98.7} & \bf{99.}6 & \bf{100}\\
    \hline	
    DeepSeg. & \bf{78.2} & \bf{94.}1 & 97.1 & 98.6 & 99.1 & 99.4\\
    \hline
    \hline
  \end{tabular}
\end{table}

\subsubsection{Results --- \textsc{BB}}
For the \textsc{bb} dataset we use two layers of LSTMs with dropout after each recurrent layer. We have experiences with bidirectional LSTMs as well but only forward LSTM performs better on this dataset. We use \eq{eq:loss} as our loss function. We use the same features as in \cite{sonderegger2012automatic, adiautomatic}, overall we have 51 features per frame. We optimize the networks using AdaGrad optimization. All parameters were tuned on a dedicated development set. Table ~\ref{tab:BBLoss} summarize the results using the loss function as in \cite{sonderegger2012automatic}. It is worth notice that we see the same behavior on this dataset as well, regarding the DeepVOT preforms better then the \segmentor in hight tolerance values.

\begin{table}[!h]
\small
\renewcommand{\arraystretch}{1.5}
	\centering
	\caption{\it Proportion of differences between automatic and manual measures falling at or below a given tolerance value (in msec). These results are for the \textsc{bb} dataset.} 
  \vspace{5pt}
  \label{tab:BBLoss}  
  \begin{tabular}{lcccccc}
    \hline
    \hline
    Model & $t\le$2 & $t\le$5 & $t\le$10 & $t\le$15 & $t\le$25 & $t\le$50\\
    \hline    
    AutoVOT& 59.1 & 80.5  & 89.9 & 94.3 & 96.8 & 98.1\\    
    \hline
    DeepVOT & 60.3 & 84.2  & \bf{94.3} & 94.9 & \bf{98.1} & \bf{98.7}\\
    \hline
    DeepSeg. & \bf{64.8} & \bf{85.5} & \bf{94.3} & \bf{95.0} & 96.2 & 97.5\\
    \hline
    \hline
  \end{tabular}
\end{table}

\label{sec:disc}
\section{Future work}
Future work will explore timing sequence of length greater than 2 - for instance, in phoneme segmentation, where the sequence varies across training examples. The model's robustness to noise and length as well as its ability to generalize are also key areas of future development. We would therefore like to explore training the model in two stages: first as a multi-class version and then fine-tuning using structured loss. With respect to machine learning, future directions include the effect of network size, depth, and loss function on model performance.

\label{sec:con}
In this paper we present a new algorithm for speech segmentation and evaluate its performance to two different tasks. The proposed algorithm combines structured loss function with recurrent neural networks and outperforms current state-of- the-art methods.

\bibliographystyle{IEEEbib}
\bibliography{bib}

\end{document}